\documentclass[a4paper,11pt]{article}

\usepackage{fourier}
\usepackage{color}
 \usepackage{graphicx}
\usepackage{url}
\usepackage[affil-it]{authblk}
\usepackage{amsmath}
\usepackage{wrapfig}

\usepackage[T1]{fontenc}
\usepackage{times}

\begin{document}
\title{Will your Doorbell Camera still recognize you as you grow old?}

\author[1]{Wang Yao} 
\author[1]{Muhammad Ali Farooq}
\author[2]{Joseph Lemley}
\author[1]{Peter Corcoran}

\affil[1]{School of Engineering, University of Galway, Ireland.}
\affil[2]{Xperi Corporation, Galway.}
\date{}
\maketitle
\thispagestyle{empty}

\vspace{-30pt}
\begin{abstract}
Robust authentication for low-power consumer devices such as doorbell cameras poses a valuable and unique challenge. This work explores the effect of age and aging on the performance of facial authentication methods.
Two public age datasets, AgeDB and Morph-II have been used as baselines in this work. A photo-realistic age transformation method has been employed to augment a set of high-quality facial images with various age effects. Then the effect of these synthetic aging data on the high-performance deep-learning-based face recognition model is quantified by using various metrics including Receiver Operating Characteristic (ROC) curves and match score distributions. 
Experimental results demonstrate that long-term age effects are still a significant challenge for the state-of-the-art facial authentication method.
\end{abstract}

\section{Introduction}
\begin{wrapfigure}{r}{0.45\textwidth}
  \vspace{-20pt}
  \begin{center}
    \includegraphics[width=0.43\textwidth]{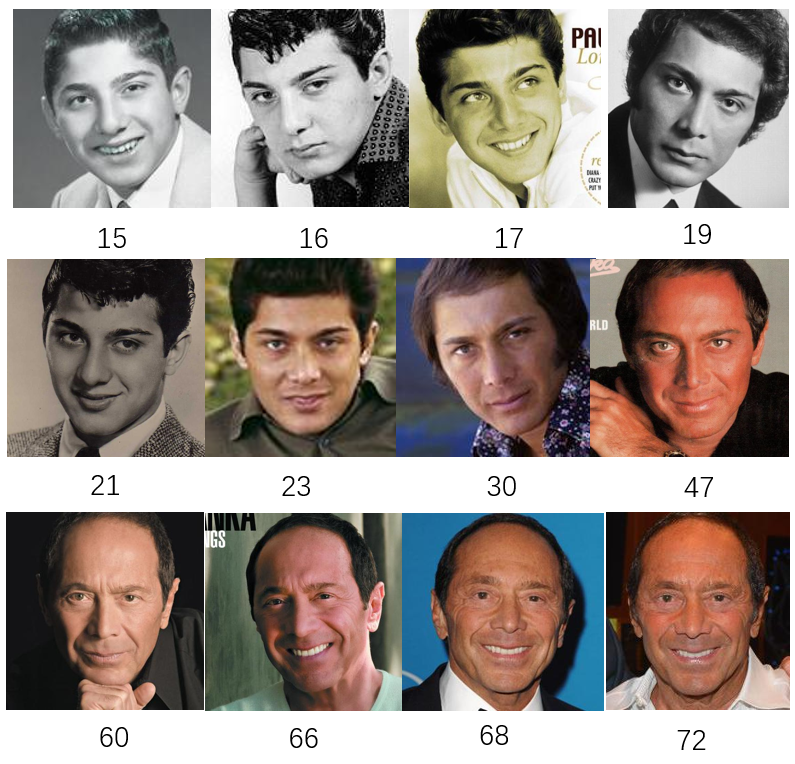}
\end{center}
\vspace{-20pt}
\caption{Example of an individual facial variation during the aging process (Images from~\cite{Moschoglou_2017_CVPR_Workshops}).}
\label{RealSample}
  \vspace{-10pt}
\end{wrapfigure}
Automatic face authentication/recognition is one of the active and long-standing research topics in the field of computer vision. 
Studies~\cite{TASKIRAN2020102809} have shown that factors such as age, lighting, and pose can have a significant impact on the performance of face recognition algorithms. 
Recent work~\cite{yao2022toward} has studied the main factors such as illumination and pose that affect facial authentication, which shows the feasibility of implementing a robust authentication method for low-power consumer devices. In this work, we are focusing on another challenging factor, age, i.e., how age bias affects the state-of-the-art Face Recognition (FR) method.

Age progression as a basic demographic can be used for real-time biometric authentication applications such as doorbell cameras and FaceID for cell phones.
It is necessary to obtain sufficient aging data from diverse identities to study the robustness of biometric authentication systems to age effects. 
Figure~\ref{RealSample} shows an example of the aging effects of a celebrity. 
However, collecting data on a person from birth to old age is difficult, and existing age datasets rarely have large amounts of face data from multiple identities at different stages of age. In this work, a generative adversarial network (GAN) based aging technique has been used to create a synthetic aging dataset, which could generate different target ages of an identity. 
In this way, the robustness of the FR model to aging can be verified without a need to collect data from human subjects over long periods of time.
The main contributions of this paper are as follows.
\begin{itemize}
    \item[(1)] This work quantifies the long-term aging effect on state-of-the-art neural face recognition algorithm.
    \vspace{-8pt}
    \item[(2)] This work explores the feasibility of using synthetic aging data to augment real-world age data.
    \vspace{-8pt}
    \item[(3)] Experimental results show that large age intervals cause obvious degradation in the FR algorithm.
\end{itemize}

\section{Background}
The development of face authentication technology in recent years has resulted in the emergence of many state-of-the-art FR algorithms based on deep learning. These algorithms have achieved remarkable performance improvements on large-scale data sets by optimizing deep neural networks and loss functions.
Studies have shown that the performance of face recognition is affected by demographics including age, race, and gender which can produce bias. 
For example, \cite{deb2017face} present a longitudinal study of FR and find a significant decrease in the accuracy of COTS-A and COTS-B face matchers over a time interval of 8.5 years. \cite{9671213} find that the recognition rate becomes significantly low when age intervals are greater than 20 years.
These studies always adopted feature extraction-based FR methods.
This raises the following question in this study. How robust is the recent deep learning-based FR method to age bias?

\section{Methodology}

\begin{figure*}[!ht]
    \centering
    \includegraphics[width=0.7\linewidth]{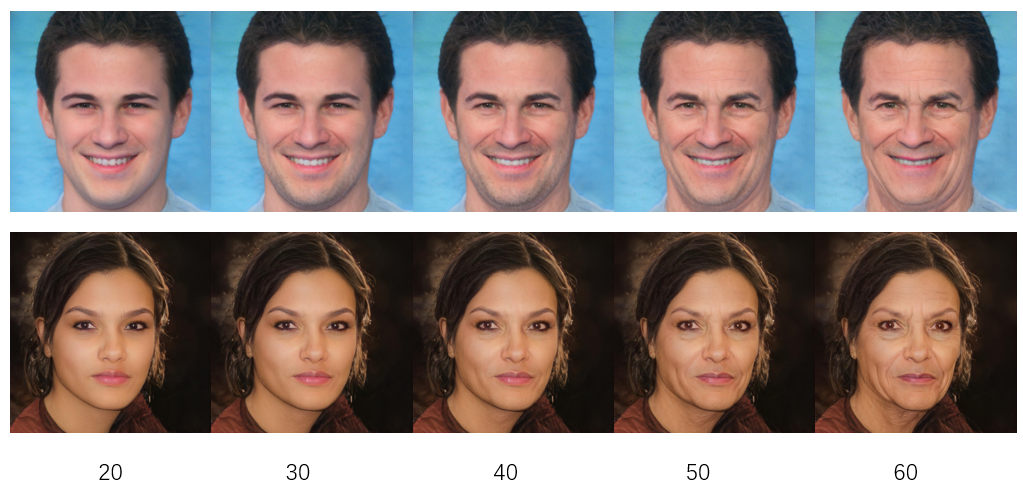}
    \vspace{-10pt}
    \caption{Samples of age variations using the synthetic aging method.}
    \vspace{-20pt}
    \label{SyntheticExample}
\end{figure*}

\paragraph{Datasets:}
Two publicly available age datasets AgeDB and Morph II are adopted in this work to study the age effect on the FR algorithm.
\noindent
(a) AgeDB~\cite{Moschoglou_2017_CVPR_Workshops} contains 568 subjects with 16488 images. The average number of images per subject is 29. 
\noindent
(b) Morph II~\cite{Ricanek2006} contains 55134 images with 13618 identities taken from 2003 to 2007. Each subject has an average of 4 images. 
\vspace{-10pt}
\paragraph{Synthetic Aging Method:}
SAM~\cite{Alaluf2021}, as one of the state-of-the-art synthetic aging techniques is adopted in this work to generate more aging samples. SAM treats the aging process as an image-to-image transformation problem by pairing a pre-trained fixed StyleGAN generator with an encoder. The task of the encoder is to encode real face images directly into the StyleGAN latent space to obtain the faces with expected age change.
Some synthetic aging samples are shown in Figure~\ref{SyntheticExample}.

\paragraph{Face Recognition Method:}
ArcFace~\cite{deng2019arcface} is selected as the FR model to verify the validity of synthetic aging faces, which uses additive angular margin loss to obtain good intra-class compactness and inter-class dispersion. In this work, MTCNN 
is employed to detect and crop the faces. Then, these faces are fed into the FR network to obtain facial features. The identity similarity score is derived by calculating the cosine similarity of the two facial features. 

\vspace{-10pt}
\paragraph{Evaluation Metrics:}
Two evaluation metrics including the receiver operating characteristic (ROC) curve and match score distribution have been adopted in our experiment to quantify the effect of synthetic aging data on the high-performance deep-learning-based FR model.

\section{Experiment}
\begin{wrapfigure}{r}{0.5\textwidth}
  \vspace{-20pt}
  \begin{center}
    \includegraphics[width=0.45\textwidth]{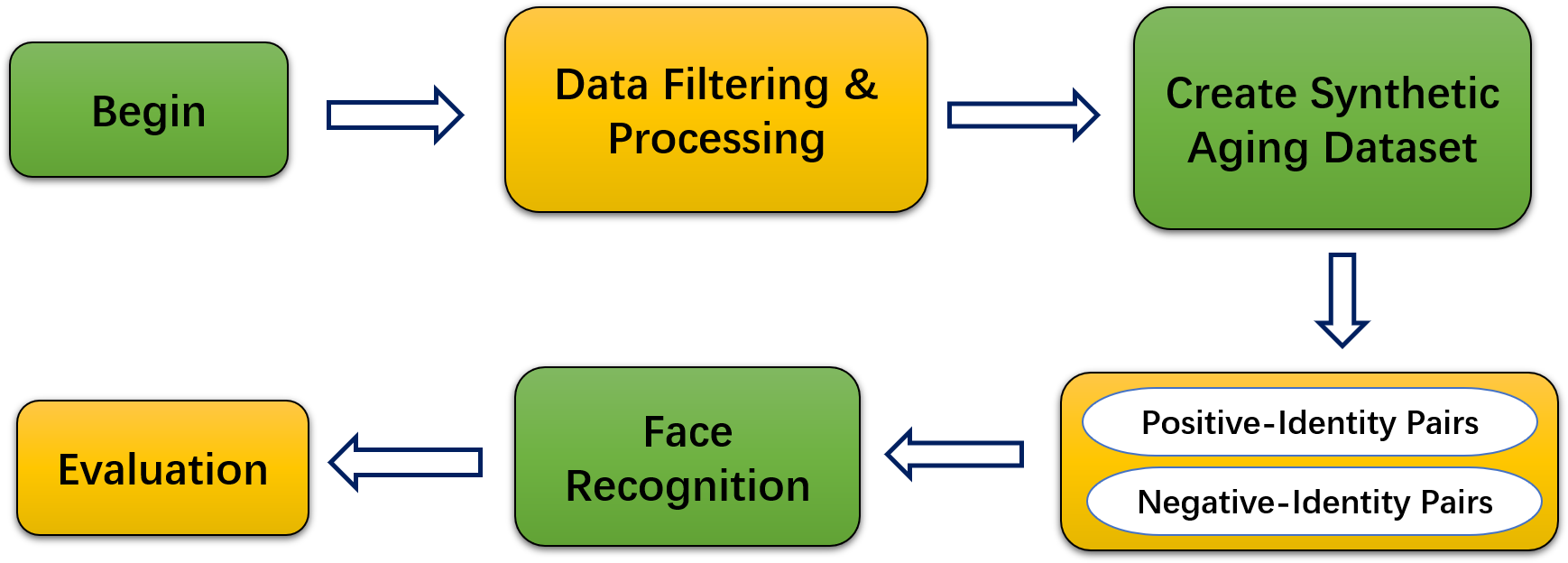} 
    \end{center}
\vspace{-20pt}
\caption{Experimental Process.}
\label{process}
  \vspace{-10pt}
\end{wrapfigure}
\paragraph{Experimental Setting:} The process of this experiment is shown in Figure~\ref{process}. First, all detectable face images from 20 to 30 years old are selected for pre-processing to obtain $256 \times 256$ images. Then, these face images are taken as the original images to synthesize faces of different ages, and here we synthesize face images of 20, 30, 40, 50, 60, 70, and 80 years old respectively. Moreover, positive-identity pairs (PPs) and negative-identity pairs (NPs) are formed by using the original images. A positive-identity pair means an image pair from the same identity. A negative-identity pair means an image pair from different identities. For different age intervals, one of the images from PPs/NPs is replaced separately with the images of the target age. Finally, the images from PPs/NPs are fed into the FR model to calculate the similarity scores, which are used for evaluation.
\subsection{ROC Comparison}
\begin{wrapfigure}{r}{0.5\textwidth}
  \vspace{-50pt}
  \begin{center}
    \includegraphics[width=0.45\textwidth]{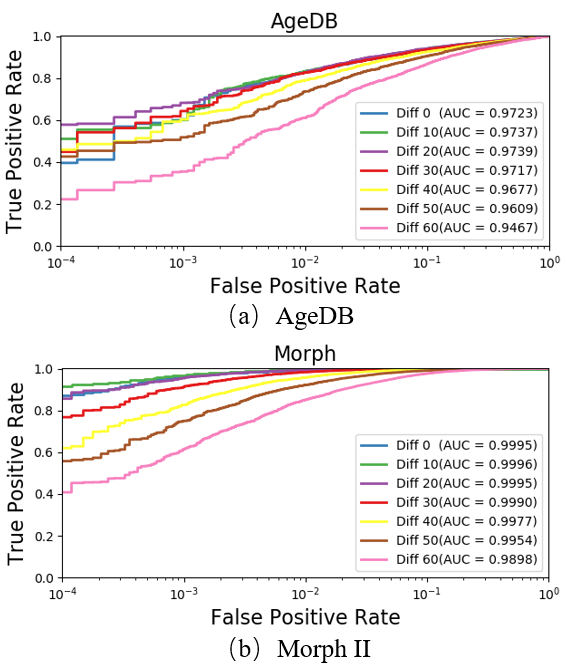} 
\end{center}
\vspace{-20pt}
\caption{The ROC curves for ArcFace of AgeDB and Morph II.}
\label{ROC}
  \vspace{-30pt}
\end{wrapfigure}
The ROC curves in Figure~\ref{ROC} quantify the effect of different age intervals on the FR algorithm.
Age intervals within 20 years have a weak effect on the performance of the state-of-the-art FR model. There is a slight degradation of the FR model performance for a 30-year time interval.
Longer time intervals still cause significant degradation in the performance of the FR model.

\subsection{Match Score Comparison}
\begin{figure*}[!ht]
    \centering
    \includegraphics[width=0.8\linewidth]{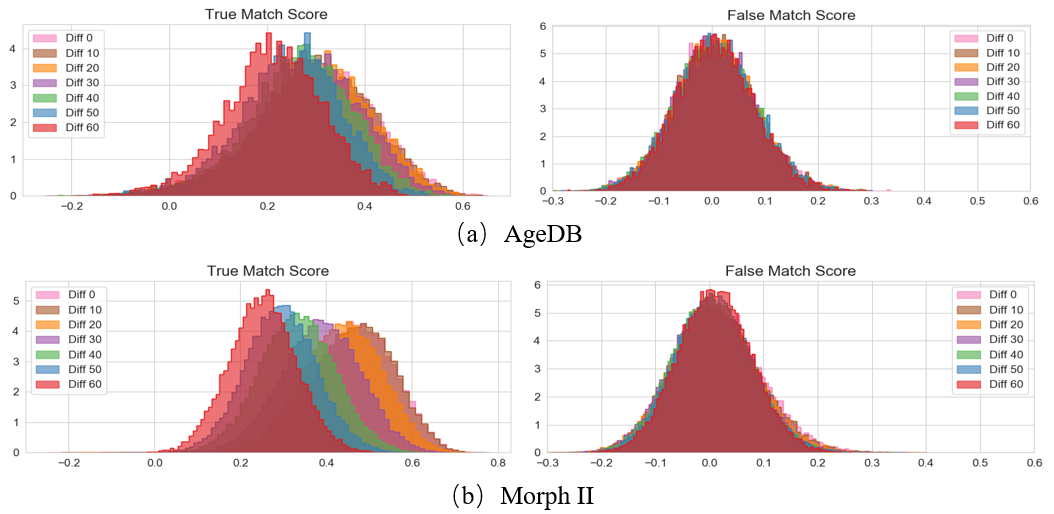}
    \vspace{-10pt}    
    \caption{The match score distributions for ArcFace of AgeDB and Morph II.}
    \vspace{-10pt}
    \label{Match}
\end{figure*}

Figure~\ref{Match} shows the impostor and genuine distributions of different age intervals. The impostor distributions 
are similar across different age intervals, while the genuine distributions with a noticeable shift to the left as the age intervals get larger. As the age interval increases, the area where the impostor distribution intersects with the true distribution increases, and the recognition performance decreases. This indicates that the genuine distribution is the main reason leading to the degradation of the performance of the FR algorithm.

\section{Conclusion}
This work qualifies the impact of age intervals on a SOTA FR algorithm and illustrates the potential value of synthetic age data in analyzing the robustness of face authentication systems. Initial experiments associated with synthetic age data have shown that it is valid to use synthetic data to enlarge the dataset for analysis of how aging affects FR models. The evaluation results from ROC curves and match score comparison show that age differences within 20 years do not have a noticeable impact on face recognition accuracy, while long-term age differences remain a significant challenge for the current facial authentication method. 
In future work, we will explore the quality of synthetic age data, and do a comparison of the real age data and the synthetic age data. We plan to expand this work to create a synthetic age dataset and to implement a privacy-secured robust face authentication solution on low-power neural accelerators.

\section*{Acknowledgments}

This research is supported by (i) Irish Research Council Enterprise Partnership Ph.D. Scheme (Project ID: EPSPG/2020/40), (ii) Xperi Corporation, Ireland, and (iii) the Data-Center Audio/Visual Intelligence on-Device (DAVID) Project (2020–2023) funded by the  Disruptive Technologies Innovation Fund (DTIF).


\appendix

\bibliographystyle{apalike}

\bibliography{imvip}

\begin{thebibliography}{}

\bibitem[Alaluf et~al., 2021]{Alaluf2021}
Alaluf, Y., Patashnik, O., and Cohen-Or, D. (2021).
\newblock Only a matter of style: Age transformation using a style-based
  regression model.
\newblock {\em ACM Trans. Graph.}, 40(4).

\bibitem[Boussaad and Boucetta, 2021]{9671213}
Boussaad, L. and Boucetta, A. (2021).
\newblock The aging effects on face recognition algorithms: the accuracy
  according to age groups and age gaps.
\newblock In {\em 2021 International Conference on Artificial Intelligence for
  Cyber Security Systems and Privacy (AI-CSP)}, pages 1--6.

\bibitem[Deb et~al., 2017]{deb2017face}
Deb, D., Best-Rowden, L., and Jain, A.~K. (2017).
\newblock Face recognition performance under aging.
\newblock In {\em Proceedings of the IEEE Conference on Computer Vision and
  Pattern Recognition Workshops}, pages 46--54.

\bibitem[Deng et~al., 2019]{deng2019arcface}
Deng, J., Guo, J., Xue, N., and Zafeiriou, S. (2019).
\newblock Arcface: Additive angular margin loss for deep face recognition.
\newblock In {\em Proceedings of the IEEE/CVF conference on computer vision and
  pattern recognition}, pages 4690--4699.

\bibitem[Moschoglou et~al., 2017]{Moschoglou_2017_CVPR_Workshops}
Moschoglou, S., Papaioannou, A., Sagonas, C., Deng, J., Kotsia, I., and
  Zafeiriou, S. (2017).
\newblock Agedb: The first manually collected, in-the-wild age database.
\newblock In {\em Proceedings of the IEEE Conference on Computer Vision and
  Pattern Recognition (CVPR) Workshops}.

\bibitem[Ricanek and Tesafaye, 2006]{Ricanek2006}
Ricanek, K. and Tesafaye, T. (2006).
\newblock Morph: a longitudinal image database of normal adult age-progression.
\newblock In {\em 7th International Conference on Automatic Face and Gesture
  Recognition (FGR06)}, pages 341--345.

\bibitem[Taskiran et~al., 2020]{TASKIRAN2020102809}
Taskiran, M., Kahraman, N., and Erdem, C.~E. (2020).
\newblock Face recognition: Past, present and future (a review).
\newblock {\em Digital Signal Processing}, 106:102809.

\bibitem[Yao et~al., 2022]{yao2022toward}
Yao, W., Varkarakis, V., Costache, G., Lemley, J., and Corcoran, P. (2022).
\newblock Toward robust facial authentication for low-power edge-ai consumer
  devices.
\newblock {\em IEEE Access}, 10:123661--123678.

\end{thebibliography}

\end{document}